\documentclass[runningheads]{llncs}
\usepackage{booktabs} 
\usepackage{amsmath,url,graphicx,amssymb}
\usepackage{amsfonts}
\usepackage{multirow}
\usepackage{algorithm}
\usepackage{algpseudocode}
\usepackage{pifont}
\usepackage{color}
\usepackage{xspace}
\usepackage{enumitem}
\usepackage{subcaption}
\usepackage{cite}
\usepackage{colortbl}
\usepackage{xcolor}
\usepackage{bm}
\usepackage{caption} 
\usepackage[
  separate-uncertainty = true,
  multi-part-units = repeat
]{siunitx}

\newcommand{\mylistbegin}{
  \begin{list}{$\bullet$}
   {
     \setlength{\itemsep}{-2pt}
     \setlength{\leftmargin}{1em}
     \setlength{\labelwidth}{1em}
     \setlength{\labelsep}{0.5em} } }
\newcommand{\mylistend}{
   \end{list}  }

\newcommand{\eg}{\textit{e.g.}}

\newcommand{\ie}{\textit{i.e.}}

\newcommand{\header}[1]{{\vspace{+1mm}\flushleft \textbf{#1}}}

\usepackage{graphicx}
\begin{document}
\newcommand{\themodel}{COACHER\xspace}

\setlength{\floatsep}{4pt plus 4pt minus 1pt}
\setlength{\textfloatsep}{4pt plus 2pt minus 2pt}
\setlength{\intextsep}{4pt plus 2pt minus 2pt}
\setlength{\dbltextfloatsep}{3pt plus 2pt minus 1pt}
\setlength{\dblfloatsep}{3pt plus 2pt minus 1pt} 

\setlength{\abovecaptionskip}{3pt}
\setlength{\belowcaptionskip}{2pt}
\setlength{\abovedisplayskip}{2pt plus 1pt minus 1pt}
\setlength{\belowdisplayskip}{2pt plus 1pt minus 1pt}

\title{Zero-Shot Scene Graph Relation Prediction through Commonsense Knowledge Integration}
\titlerunning{COACHER}

%
\author{Xuan Kan \and Hejie Cui \and Carl Yang\thanks{Corresponding Author.}}
\authorrunning{X. Kan et al.}
%
\institute{Department of Computer Science, Emory University \\
\email{\{xuan.kan, hejie.cui, j.carlyang\}@emory.edu}
}

\maketitle              
\begin{abstract}
Relation prediction among entities in images is an important step in scene graph generation (SGG), which further impacts various visual understanding and reasoning tasks. Existing SGG frameworks, however, require heavy training yet are incapable of modeling unseen (\ie, zero-shot) triplets. 
In this work, we stress that such incapability is due to the lack of commonsense reasoning, \ie, the ability to associate similar entities and infer similar relations based on general understanding of the world. To fill this gap, we propose \textbf{C}omm\textbf{O}nsense-integr\textbf{A}ted s\textbf{C}ene grap\textbf{H} r\textbf{E}lation p\textbf{R}ediction (\textbf{COACHER}), a framework to integrate commonsense knowledge for SGG, especially for zero-shot relation prediction. Specifically, we develop novel graph mining pipelines to model the neighborhoods and paths around entities in an external commonsense knowledge graph, and integrate them on top of state-of-the-art SGG frameworks. Extensive quantitative evaluations and qualitative case studies on both original and manipulated datasets from Visual Genome demonstrate the effectiveness of our proposed approach. The code is available at \url{https://github.com/Wayfear/Coacher}.

\end{abstract}

\section{Introduction}
\label{sec:intro}
With the unprecedented advances of computer vision, visual understanding and reasoning tasks such as Image Captioning and Visual Question Answer (VQA) have attracted increasing interest recently. Scene graph generation (SGG), which predicts all relations between detected entities from images, distills visual information in structural understanding. With clear semantics of entities and relations, scene graphs are widely used for various downstream tasks like VQA \cite{teney2017graph, hudson2019gqa, zhu2015building}, image captioning \cite{yao2018exploring, yang2019auto, gu2018stack}, and image generation \cite{li2020lightweight, li2020manigan, li2019controllable}.

A critical and challenging step in SGG is \textit{relation prediction}. A relation instance in scene graph is defined as a triplet $\langle$\textit{subject, relation, object}$\rangle$. Given two detected entities, which relation exists between them is predicted based on the probability score from the learned relation prediction model. However, most of the existing scene graph generation models rely on heavy training to memorize seen triplets, which limits their utility in reality, because many real triples are never seen during training. \\

\begin{figure*}[t]
    \centering
    \includegraphics[width=\linewidth]{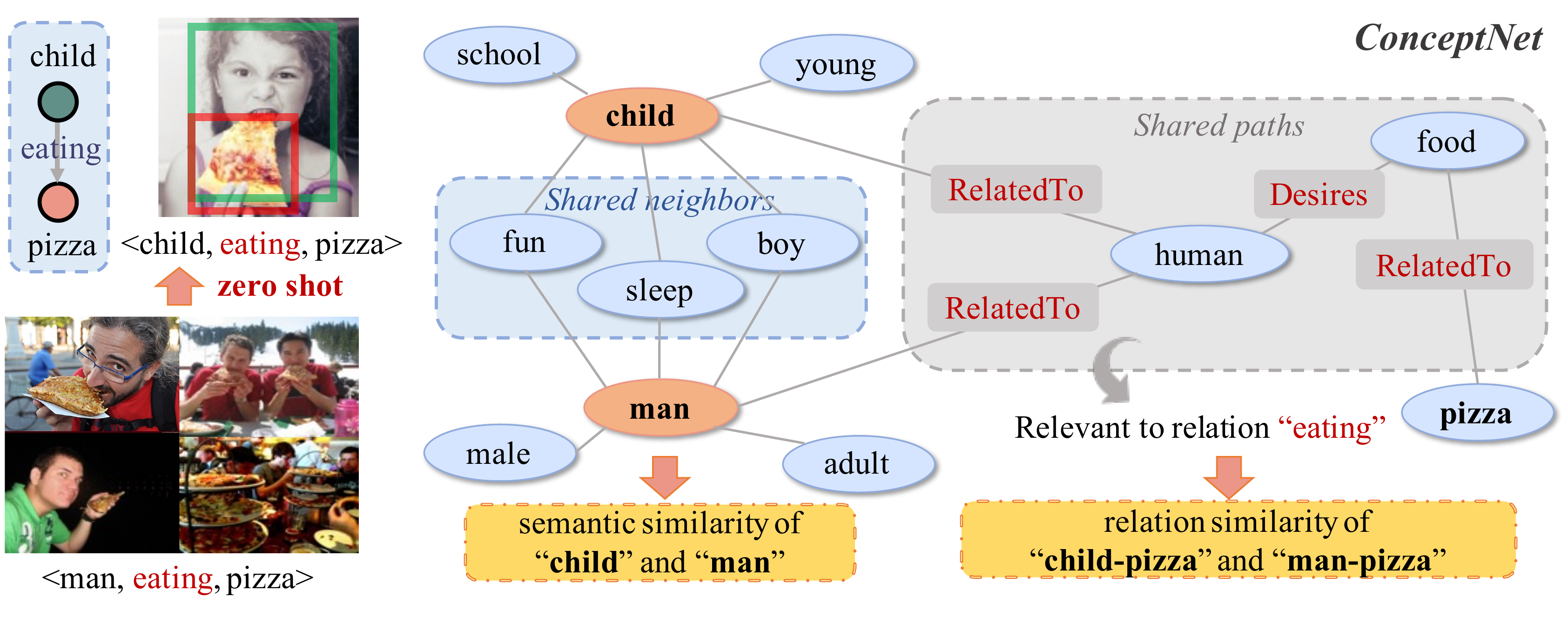}
    \caption{Toy example of zero-shot relation prediction in scene graph generation and insights about commonsense knowledge integration. 
    }
    \label{fig:toy-example}
\end{figure*}

\noindent \textbf{Challenge: Performance deterioration on zero-shot triplets.} When evaluating the performance of relation prediction model in practice, there are two types of triplets, the ones seen in the training data and the ones unseen. Those unseen ones are called \textit{zero-shot} triplets. As shown in Figure \ref{fig:toy-example}, triplet $\langle$\textit{man, eating, pizza}$\rangle$ is observed in the training data. If this triplet appears again in the testing phase, then it is called a non-zero-shot triplet. In contrast, a triplet $\langle$\textit{child, eating, pizza}$\rangle$ with a new entity-relation combination not observed in training data is called a zero-shot triplet. 

Although several existing scene graph generation methods have achieved decent performance on the whole testing data, little analysis has been done on the performance on zero-shot triplets. Unfortunately, based on our preliminary analysis (Section 2.1), the performance of existing methods degrades remarkably when solely tested on zero-shot triplets, while many of such triplets are rather common in real life. Such reliance on seeing all triplets in training data is problematic, as possible triplets in the wild are simply inexhaustible, which requires smarter models with better generalizability. 

\noindent \textbf{Motivation: Commonsense as a ``coacher'' for zero-shot relation prediction.}
Commonsense knowledge refers to general facts about the world that empower human beings to reason over unfamiliar scenarios. Motivated by this process from humans' perspective, in this work, we propose to integrate commonsense knowledge to alleviate the inexhaustible-triplet problem and improve the performance of zero-shot relation prediction in SGG. 

The illustration of our problem and insights are illustrated in Figure \ref{fig:toy-example}. 
Specifically, the commonsense knowledge utilized in this paper comes from ConceptNet \cite{ConceptNet}, a crowd-sourced semantic knowledge graph containing rich structured knowledge regarding real-world concepts.

\noindent \textbf{Insight 1: Neighbor commonsense reflects semantic similarity.}
In ConceptNet, the neighbor similarity between two individual nodes indicates their semantic similarity in the real world. For example, in Figure \ref{fig:toy-example}, \textit{child} and \textit{man} share many common neighbors such as \textit{fun}, \textit{sleep}, \textit{boy} and so on, which indicates that \textit{child} and \textit{man} may be similar and thus have similar interactions with other entities. If the model sees a triplet $\langle$\textit{man, eating, pizza}$\rangle$ in the training data, then with the knowledge that \textit{child} is semantically similar to \textit{man}, it should more easily recognize triplets like $\langle$\textit{child, eating, pizza}$\rangle$ from unseen but similar images. 
Therefore, we propose to leverage the semantic similarity between two detected entities by modeling their neighborhood overlap in ConceptNet. \\

\noindent \textbf{Insight 2: Path commonsense reflects relation similarity.}
Nodes are connected by paths composed of multiple consecutive edges in ConceptNet. As is shown on the right in Figure \ref{fig:toy-example}, the entity pairs of $(child, pizza)$ and $(man, pizza)$ share common intermediate paths like $\langle$\textit{RelatedTo, human, Desires, food, RelatedTo}$\rangle$. This similarity of intermediate paths indicates that the relations between \textit{man} and \textit{pizza} may be similar to those between \textit{child} and \textit{pizza}. If there is a triplet $\langle$\textit{man, eating, pizza}$\rangle$ in the training data, then the model should tend to predict the relation $eating$ given (\textit{child}, \textit{pizza}) in an unseen but similar image.
Following the idea above, we propose to infer the relation between two entities by modeling their path coincidence with other entity pairs in ConceptNet. \\

\noindent \textbf{Approach: Scene Graph Relation Prediction through Commonsense Knowledge Integration.}
\noindent In this work, we propose a novel framework that integrates external commonsense knowledge into SGG for relation prediction on zero-shot triplets, which we term as  \underline{C}omm\underline{O}nsense-integr\underline{A}ted s\underline{C}ene grap\underline{H} r\underline{E}lation  p\underline{R}ediction, \themodel for brevity (Figure \ref{fig:method}). To be specific, we investigate the utility of external commonsense knowledge through real data analysis. With the validated effectiveness of both neighbor- and path-based commonsense knowledge from ConceptNet, we design three modules for different levels of knowledge integration, which generate auxiliary knowledge embedding for generalizable relation prediction. 

In summary, our main contributions are three-fold.
\begin{itemize}
	\item [$\bullet$] We analyze the ignorance of zero-shot triplets by existing SGG models and validate the potential utility of commonsense knowledge from ConceptNet through real data analysis (Section \ref{sec:motivate}).
	\item [$\bullet$] Based on the state-of-the-art SGG framework, we integrate external commonsense knowledge regarding ConceptNet neighbors and paths to improve relation prediction on zero-shot triplets (Section \ref{sec:method}).
	\item [$\bullet$] Extensive quantitative experiments and qualitative analyses on the widely-used SGG benchmark dataset of Visual Genome demonstrate the effectiveness of our proposed \themodel framework. Particularly, \themodel achieves consistently better performance over state-of-the-art baselines on the original dataset, and outperforms them significantly on amplified datasets towards more severe zero-shot learning settings (Section \ref{sec:exp}).
\end{itemize}

\section{Motivating Analysis}
\label{sec:motivate}

\subsection{Ignorance yet Importance of Zero-Shot Triplets}
In SGG, triplets are used to model entities with their relations. Among the great number of possible rational triplets in the wild, some of them exist in the training data while more others do not. The ability of correctly inferring zero-shot triplets can be extremely important to reflect the generalization capability of the model and its real utility in practice. 

Although the performance of zero-shot scene graph generation was once studied in the early days on a small dataset \cite{lu2016visual}, later researchers do not pay much attention to this setting. Until 2020, Tang et al.~\cite{UnbiasedTang_2020} first reported zero-shot performance on Visual Genome. However, they have not proposed any particular solutions to improve it. 

\begin{table}
\centering
\small
\caption{Performance (\%) of three state-of-the-art models on non-zero-shot and zero-shot triplets on Visual Genome (Please refer to Section 4 for more details about the presented models).}
\label{tab:zero_no_zeroperformance}
\resizebox{0.8\textwidth}{!}{
\begin{tabular}{cccccccccccc}
\toprule
{Methods} &\multicolumn{3}{c}{\bf NM} & & \multicolumn{3}{c}{\bf NM+}& & \multicolumn{3}{c}{\bf TDE}\\
\cmidrule(lr){2-4} \cmidrule(lr){6-8} \cmidrule(lr){10-12} 
{MeanRecall@\it K} & {\it K=$20$} & {\it K=$50$} & {\it K=$100$} & { } & {\it K=$20$} & {\it K=$50$} & {\it K=$100$} & { } & {\it K=$20$} & {\it K=$50$} & {\it K=$100$}\\
\midrule
{\bf none-zero-shot} & 25.12 & 33.32 & 37.06 & & 25.08 & 33.69 & 37.54 & & 26.26 & 35.93 & 40.27 \\
\rowcolor{gray!20} {\bf zero-shot} &  12.85 & 18.93 & 21.84 & & 12.28 & 18.28 & 21.30 & & 5.84 & 11.68 & 15.10 \\
\bottomrule
\end{tabular}}
\end{table}

The ignorance of zero-shot settings causes existing methods a dramatic descent on the relation prediction on zero-shot triplets. Table \ref{tab:zero_no_zeroperformance} shows the performance of three state-of-the-art models on Visual Genome, the most widely used benchmark dataset for SGG. Note that \textit{mean recall} is used here for performance evaluation, which is the average result of \textit{triplet-wise} recall. Consistently, the mean recall of non-zero-shot triplets under different values of $K$ can achieve almost twice of that on zero-shot ones, which demonstrates a concerning performance deterioration on zero-shot relation prediction.

\begin{figure}
    \centering
    \includegraphics[width=0.8\linewidth]{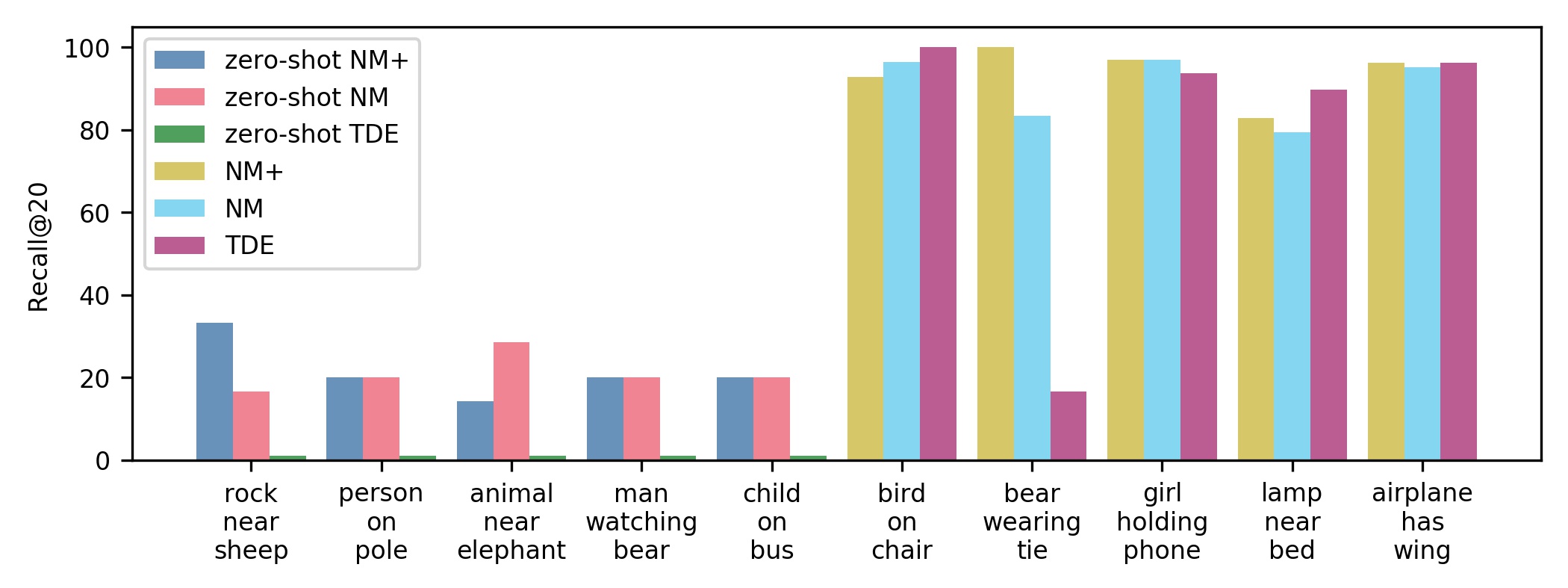}
    \caption{Recall@20 performance on two different types of triplets. The five triplets on the left are zero-shot ones while the other five on the right are non-zero-shot. }
    \label{fig:fig4}
\end{figure}
Furthermore, although zero-shot triplets are not labeled in the training data, some of them are actually no less common in reality compared to the labeled ones. To give a more concrete illustration, we visualize the \textit{recall@20} on five actual zero-shot triplets and another five non-zero-shot triplets in Figure \ref{fig:fig4}. Consistent with the results in Table \ref{tab:zero_no_zeroperformance}, performance of existing models on the five zero-shot triplets turns out to be much lower than that on the five non-zero-shot triplets. However, these zero-shot triplets represent very common relations such as $\langle$\textit{child, on, bus}$\rangle$ which are in fact more common than some non-zero-shot ones such as $\langle$\textit{bear, wearing, tie}$\rangle$. The performance on certain triplets like $\langle$\textit{bear, wearing, tie}$\rangle$ is much better simply because they appear in the training data and got memorized by the model, but the utility of such memorization is rather limited in reality without the ability of generalization.

Lastly, due to the inexhaustibility of triplets in the real world, labeled datasets can only cover a limited portion of the extensive knowledge. Many common triplets simply do not appear in the dataset. For example, $\langle$\textit{men, holding, umbrella}$\rangle$, $\langle$\textit{vehicle, parked on, beach}$\rangle$,$\langle$\textit{women, in, boat}$\rangle$ and $\langle$\textit{fence, across, sidewalk}$\rangle$, to name a few, are common triplets that never appear in Visual Genome, even though all involved objects and relations appear in other labeled triplets. Therefore, to propose a practical SGG model that can be widely used and assist various downstream tasks, we need to pay more attention to zero-shot triplets.

Motivated by the ignorance and importance of zero-shot triplets, in this work, we focus on integrating commonsense knowledge from external resources to improve the relation prediction performance on zero-shot triplets. Specifically, we leverage ConceptNet as the external knowledge resource from several other alternatives due to its wide coverage of concepts and accompanying semantic embeddings of concepts as useful features \cite{ConceptNet}. In ConceptNet, each concept (word or phrase) is modeled as a node and each edge represents the relation between two concepts. Thanks to its wide coverage, we are able to link each entity class in Visual Genome to one node in ConceptNet. 

\subsection{Commonsense Knowledge from ConceptNet Neighbors}
Many entities in real life share similar semantic meanings, which can potentially support zero-shot relation prediction. Given an image from Visual Genome, we can detect multiple entities, where each entity belongs to a class. Among these entity classes, for example, \textit{(girl, boy, woman, man, child)} are all human beings, so the model should learn to generalize human-oriented relations among them. 

The semantic similarity among classes in Visual Genome can be viewed as the neighborhood similarity of their corresponding nodes in ConceptNet, which can be calculated with the Jaccard similarity of their neighbors:
\begin{equation}
J(\mathcal{V}_A, \mathcal{V}_B)=\frac{|\mathcal{N}_A  \cap \mathcal{N}_B|}{|\mathcal{N}_A  \cup \mathcal{N}_B|} , 
\end{equation}
where $\mathcal{V}_A$, $\mathcal{V}_B$ are nodes corresponding to two given classes, and $\mathcal{N}_A$, $\mathcal{N}_B$ are the neighbors of them in ConceptNet, respectively.

\begin{table}
\centering
\caption{Top 20 pairs of similar entity classes.}
\label{tab:top20}
\resizebox{1.0\textwidth}{!}{
\begin{tabular}{cccccccccc}
\toprule
\bf 1 & \it chair-seat & \bf 2 & \it shoe-sock & \bf 3 & \it hill-mountain  & \bf 4 & \it coat-jacket & \bf 5 & \it house-building \\ 
\hline
\bf 6 & \it airplane-plane & \bf 7 & \it woman-girl & \bf 8 & \it  desk-table &   \bf 9 & \it  window-door & \bf 10 & \it  men-man  \\ 
\hline
\bf 11 & \it shirt-jacket & \bf 12 & \it  house-room & \bf 13 & \it room-building &   \bf 14 & \it girl-boy & \bf 15 & \it plate-table \\ 
\hline
\bf 16 & \it arm-leg & \bf 17 & \it chair-table &  \bf 18 & \it tree-branch & \bf 19 & \it cow-sheep & \bf 20 & \it arm-hand \\ 
\bottomrule
\end{tabular}}
\end{table}

In order to validate the effectiveness of utilizing neighborhood similarity in ConceptNet as a measurement of semantic similarity in Visual Genome, we calculate the similarity between each pair of the top 150 mostly observed classes in Visual Genome, and rank their similarity in descending order. Results of the top 20 most similar pairs are shown in Table \ref{tab:top20}. As can be seen, the top similar pairs such as $chair-seat$ indeed capture the semantic similarity between two classes, which validates the virtue of using ConceptNet neighbors to bring in commonsense knowledge for modeling the semantic similarity among entities.

\subsection{Commonsense Knowledge from ConceptNet Paths}
In ConceptNet, besides the one-hop information from neighbors, paths composed by multiple edges can further encode rich multi-hop information. Specifically, if two pairs of entities are connected by many same paths in ConceptNet, they are more likely to share similar relations. In order to investigate such path-relation correlation between node pairs on ConceptNet, we define MidPath as follows: 
\begin{definition}[MidPath]
    Given two nodes $\mathcal{V}_A$ and $\mathcal{V}_B$ in the graph, a MidPath between $\mathcal{V}_A$ and $\mathcal{V}_B$ is defined as the sequence of all intermediate edges and nodes on a path from $\mathcal{V}_A$ to $\mathcal{V}_B$, excluding both the head and tail nodes. 
\end{definition}
For example, given a path $\langle$\textit{people, RelatedTo, automobile, AtLocation, street}$\rangle$ between nodes $people$ and $street$, the corresponding MidPath is $\langle$\textit{RelatedTo, automobile, AtLocation}$\rangle$.

\begin{table}
\centering
\caption{Top 3 related MidPaths for 10 relations.}
\label{tab:top10}
\resizebox{\textwidth}{!}{
\begin{tabular}{cccc}
\toprule
\bf Relation & \bf Top1 MidPath & \bf Top2 MidPath & \bf Top3 MidPath \\ 
\toprule
parked on & \it RelatedTo-cars-RelatedTo & \it RelatedTo-driven-ReceivesAction & \it AtLocation-automobile-RelatedTo \\
\midrule
says & \it RelatedTo & \it RelatedTo-communication\_device-RelatedTo & \it RelatedTo-command-RelatedTo \\ 
\midrule
laying on & \it RelatedTo-legs-RelatedTo & \it AtLocation & \it  RelatedTo-human-RelatedTo \\ 
\midrule
wearing & \it RelatedTo-body-RelatedTo & \it RelatedTo-dress-RelatedTo & \it RelatedTo-clothing-Desires \\ 
\midrule
against & \it AtLocation-garage-AtLocation & \it RelatedTo-wall-RelatedTo & \it RelatedTo \\ 
\midrule
sitting on & \it AtLocation & \it AtLocation-human-RelatedTo & \it RelatedTo-legs-RelatedTo \\ 
\midrule
walking in & \it RelatedTo-home-RelatedTo & \it UsedFor-children-RelatedTo & \it RelatedTo-crowd-RelatedTo \\ 
\midrule
growing on & \it RelatedTo-growth-RelatedTo & \it RelatedTo-leaves-RelatedTo & \it RelatedTo-stem-RelatedTo \\ 
\midrule
watching & \it RelatedTo-human-RelatedTo & \it RelatedTo-date-RelatedTo & \it RelatedTo-female-RelatedTo \\ 
\midrule
playing & \it RelatedTo-human-RelatedTo & \it RelatedTo & \it RelatedTo-clown-RelatedTo \\ 
\bottomrule
\end{tabular}
}
\end{table}

For each relation in Visual Genome, probability analysis is done to investigate the related MidPaths. The smallest unit in the dataset is a triplet $\langle$\textit{subject, relation, object}$\rangle$. With nodes $subject$ and $object$, we can extract a set of MidPaths $\mathcal{P}$ connecting them from ConceptNet. For each path $p_i \in \mathcal{P}$ and relation $r_j \in \mathcal{R}$, the number of co-occurrence $I(\mathcal{MP}=p_i, \mathcal{R}=r_j)$ can be counted. Following the formula below, we calculate the conditional probability of observing MidPath $p_i$ given a specific relation $r_j$:
\begin{equation}
P(\mathcal{MP}=p_i|\mathcal{R}=r_j) = \frac{I(\mathcal{MP}=p_i, \mathcal{R}=r_j)}{\sum_{p_k \in \mathcal{P}}{I(\mathcal{MP}=p_k, \mathcal{R}=r_j)}}.
\end{equation}
To eliminate random effects, the probability of observing a random MidPath $p_i$ is also calculated, as follows
\begin{equation}
P(\mathcal{MP}=p_i) = \frac{I(\mathcal{\mathcal{MP}}=p_i)}{\sum_{p_k \in \mathcal{P}}{I(\mathcal{MP}=p_k)}} ,
\end{equation}
where $I(\mathcal{MP}=p_i)$ is the occurrence number of MidPath $p_i$. Now we can measure how significant MidPath $p_i$ is given a specific relation $r_j$ by
\begin{equation}
Score(p_i, r_j)=P(\mathcal{MP}=p_i|\mathcal{R}=r_j)-P(\mathcal{MP}=p_i).
\end{equation}
The higher the score is, the more significantly MidPath $p_i$ can imply relation $r_j$. Table \ref{tab:top10} shows top three MidPaths with highest scores for 10 relations. Clearly, these top MidPaths are semantically meaningful and potentially beneficial for generalizing relation predictions among entity pairs.

\section{\themodel}
\label{sec:method}

\begin{figure}
    \centering
    \includegraphics[width=0.85\linewidth]{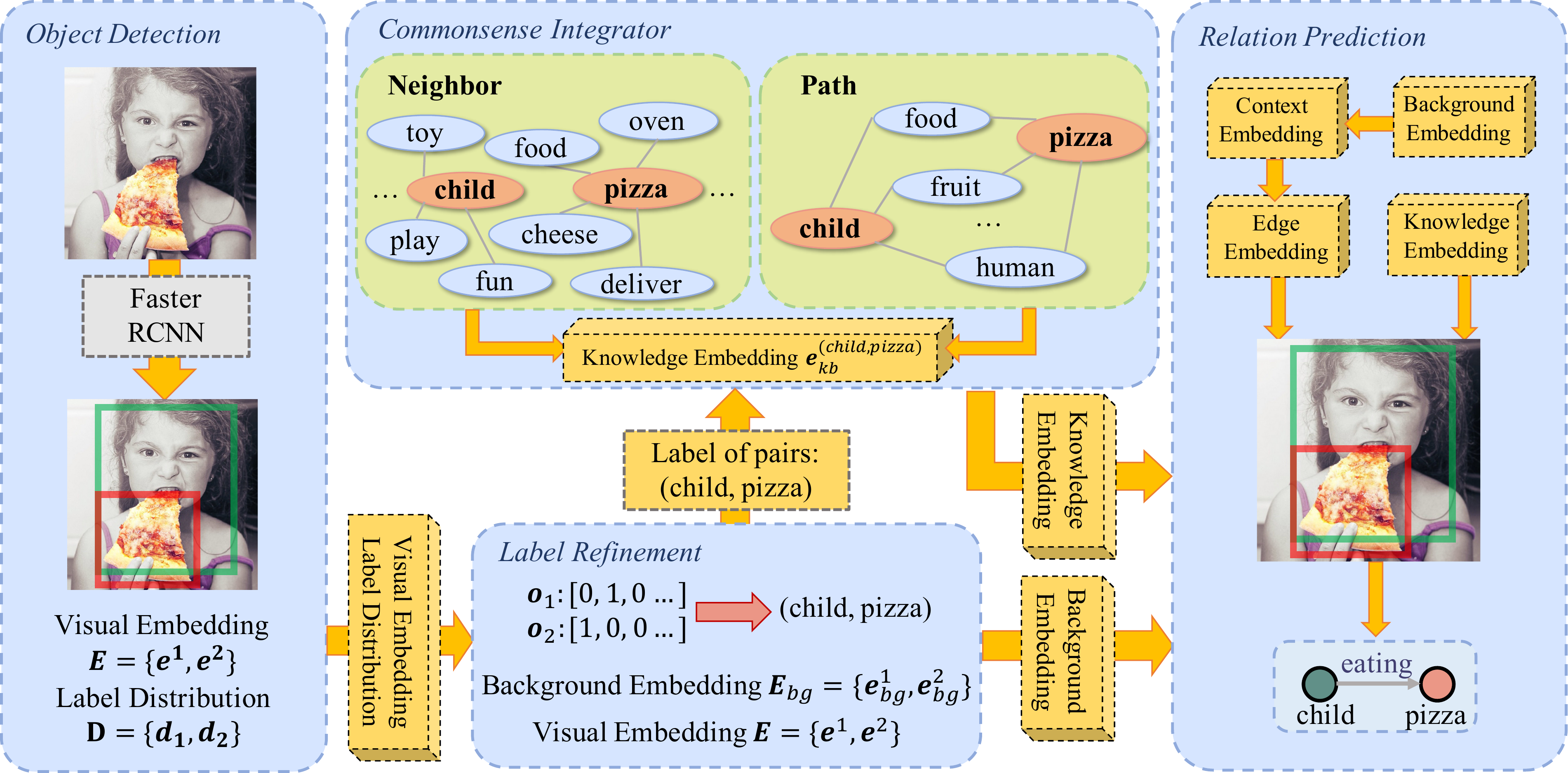}
    \caption{The overall framework of our proposed \themodel.}
    \label{fig:method}
\end{figure}

In this section, we present the proposed SGG with commonsense knowledge integration model \themodel in detail. Figure \ref{fig:method} shows the major components in \themodel: (a) object detection and refinement, (b) commonsense knowledge embedding generation, and (c) commonsense enhanced relation prediction.

\subsection{Backbone Scene Graph Generation Pipeline}
Scene graph generation is a task aiming to understand visual scenes by extracting entities from images and predicting the semantic relations between them. Various technique has been explored for scene graph generation (Section \ref{sec:related}). Here we adopt one of the state-of-the-art pipelines from Neural Motif \cite{MotifsZellers_2018} as the backbone framework. We mainly focus on commonsense-integrated relation prediction for zero-shot triplets, which is crucial for practice as discussed in Section \ref{sec:motivate}. 

The backbone scene graph generation pipeline includes three components: 
\header{Step1: object detection.} With the development of deep learning, some popular frameworks such as R-CNN \cite{girshick2015fast, ren2016faster}, YOLO \cite{yolov3} and SSD \cite{liu2016ssd} achieve impressive performance on this task. Following previous literature \cite{UnbiasedTang_2020, gu2019scene}, we adopt a pre-trained Faster R-CNN model as the detector in our framework. In this step, with the input of a single image $\mathcal{I}$, the output includes: a set of region proposals $B=\{b_1, \cdots, b_n\}$, a set of distribution vectors $\bm D=\{\bm d_1, \cdots, \bm d_n\}$, where $\bm d_i\in \mathbb{R}^{|\mathcal{C}|}$ is a label probabilities distribution and $|\mathcal{C}|$ is the number of classes, as well as a visual embedding $\bm E=\{\bm e^1, \cdots, \bm e^n\}$ for each detected object.

\header{Step2: label refinement.} 
Based on the label distribution $\bm D$ generated from Step1, we conduct label refinement to generate a one-hot vector of entity classes for each region proposal, which will be used for relation prediction.

First, background embeddings $\bm{E}_{bg}$ containing information from both region proposal level and global level of the image are generated using a bi-LSTM: 
\begin{equation}
\bm{E}_{bg} = \text{biLSTM}([\bm e^i;\text{MLP}(\bm d_i)]_{i=1,\cdots,n}), 
\end{equation}
where $\bm{E}_{bg} = [\bm{e}_{bg}^1, \bm{e}_{bg}^2, \cdots, \bm{e}_{bg}^n]$ is the hidden state of the last layer in LSTM and $n$ is the number of region proposals. Then, an LSTM is used to decode each region proposal embedding $\bm{e}_{bg}^i$:
\begin{equation}
\bm h_i = \operatorname{LSTM}([\bm{e}_{bg}^i;\bm o_{i-1}]), 
\end{equation}
\begin{equation}
\bm o_i = \operatorname{argmax}(\operatorname{MLP}(\bm h_i)).
\end{equation}
$\bm o_i$ is the one-hot vector representing the refined class label of a region proposal. 

\header{Step3: relation prediction.} After obtaining refined object labels for all region proposals, we use them to further generate context embeddings $\bm{E}_{ct}$: 
\begin{equation}
\bm{E}_{ct} = \operatorname{biLSTM}([\bm{e}_{bg}^{i}; \operatorname{MLP}(\bm{o}_i)]_{i=1,\cdots,n}),
\end{equation}
where $\bm{E}_{ct} = [\bm{e}_{ct}^1, \bm{e}_{ct}^2, \cdots, \bm{e}_{ct}^n]$, which are then used to extract edge embeddings $\bm{e}_{eg}$ and predict the relation between each pair of bounding boxes: 
\begin{equation}
\bm{e}_{eg}^{(i,j)} = \operatorname{MLP}(\bm{e}_{ct}^i) \circ \operatorname{MLP}(\bm{e}_{ct}^j) \circ (\bm e^{i} \cup \bm e^{j}),
\end{equation}
\begin{equation}
r_{(i,j)} = \operatorname{argmax}(\operatorname{MLP}([\bm{e}_{eg}^{(i,j)}; \bm{e}_{kb}^{(i,j)}])),
\end{equation}
where $\circ$ represents element-wise product, $\bm{e}_{kb}^{(i,j)}$ is the commonsense knowledge embedding obtained from ConceptNet, which we will introduce next.

\subsection{Commonsense Integrator}
Commonsense knowledge integration is achieved by computing $\bm {e}_{kb}$ from external resources. Specifically, we use ConceptNet \cite{ConceptNet} here as the source of external commonsense knowledge. ConceptNet is a knowledge graph that connects words and
phrases of natural language with labeled edges. It is constructed from rich resources such as Wiktionary and WordNet. With the combination of these resources, ConceptNet contains over 21 million edges and over 8 million nodes, covering all of the entity classes in Visual Genome. Besides, it also provides a semantic embedding for each node, which can serve as a semantic feature. Here we develop three types of integrators to generate commonsense knowledge embeddings from ConceptNet.

\header{Neighbor integrator.} ConceptNet is a massive graph $\mathcal{G} = (\mathcal{V}, \mathcal{E})$, where $\mathcal{V}$ and $\mathcal{E}$ are the node set and edge set, respectively. Each detected entity can be seen as a node in ConceptNet. Given a class $c \in \mathcal{C}$ of a detected entity, its neighborhood information can be collected as follows:
\begin{equation}
c\xrightarrow{\text{link}}\mathcal{V}_c \xrightarrow{\text{retrieve neighbors}} \mathcal{N}_{c} = \{\mathcal{V}_n|(\mathcal{V}_c, \mathcal{V}_n) \in \mathcal{E} \}.
\end{equation}
Denote $\bm{F} \subset \mathbb{R}^{|\mathcal{V}|\times k}$ as the feature matrix of all nodes in ConceptNet from \cite{ConceptNet5}, where $k$ is the dimension of the feature vector. Neighbor embedding $\bm{e}_{nb}^c$ of node $\mathcal{V}_c$ is calculated as the average over all of its neighbors' embeddings: 
\begin{equation}
\bm{e}_{nb}^c = \frac{1}{|\mathcal{N}_{c}|}\sum_{\mathcal{V}_n \in \mathcal{N}_{c}}{\bm {F}_n},
\label{eq:neiE}
\end{equation}
where $\bm {F}_n$ is the $n_{th}$ row of $\bm {F}$.
For relation prediction, given a pair of detected entities with classes $(a, b)$, the neighbor-based commonsense knowledge embedding $\bm{e}_{kb}^{(a,b)}$ of this pair is calculated as: 
\begin{equation}
\bm{e}_{kb}^{(a,b)} = \text{ReLU}(\operatorname{MLP}([\bm{e}_{nb}^a; \bm{e}_{nb}^b])),
\end{equation}
where $\operatorname{MLP}$ denotes the multi-layer perceptron.

\header{Path integrator.} Given a pair of entities with classes $(a, b)$ recognized from the object detection model, a set of paths connecting them can be obtained following the procedure below:
\begin{equation}
(a,b)\xrightarrow{\text{link}}(\mathcal{V}_a, \mathcal{V}_b) \xrightarrow{\text{retrieve paths}} \mathcal{P}_{(a, b)} = \{p|p=\{\mathcal{V}_a, \mathcal{V}_1, \cdots, \mathcal{V}_b\} \}. 
\end{equation}

\begin{figure}
    \centering
    \includegraphics[width=0.9\linewidth]{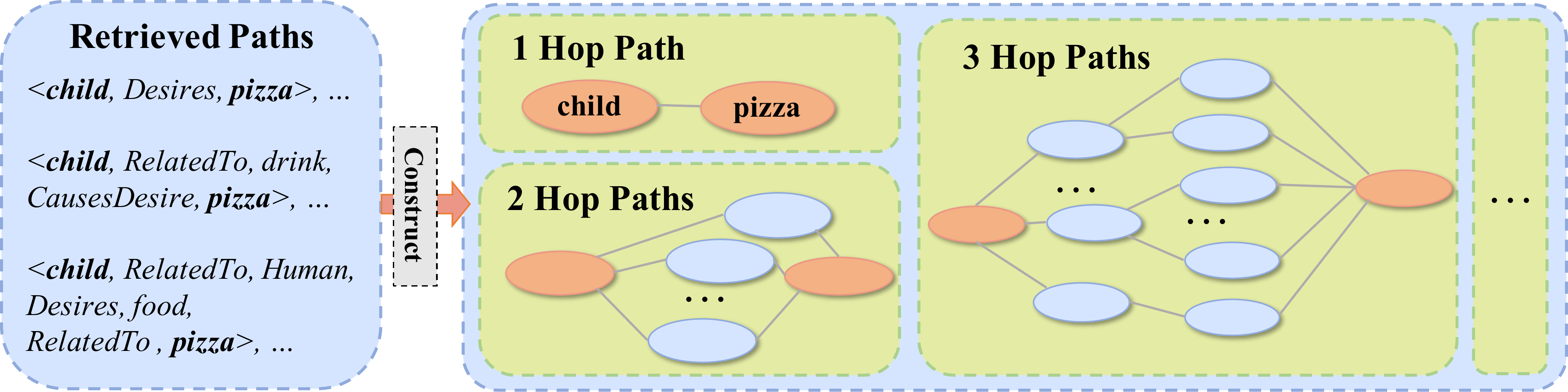}
    \caption{Graphic representation construction from retrieved paths.}
    \label{fig:retrievedpath}
\end{figure}

\noindent We classify these extracted paths based on their number of hops. The graphical representation of paths is shown in Figure \ref{fig:retrievedpath}. Each set of \textit{l-Hop Paths} naturally constitutes a small graph $G_l^{(a,b)}$, where $l$ is the number of hops on the paths. The goal is to learn a representation of all graphs $\{G_l^{(a, b)}\}_{l=1}^L$, where $L$ is often a small number like 2 or 3, 
since longer paths are too noisy and hard to retrieve. 

Classical sequence models such as LSTM cannot handle very short paths effectively. 
Inspired by message passing network for graph representation learning \cite{murphy1999loopy, gilmer2017neural}, we design a neural message passing mechanism to learn a representation for each set of \textit{l-Hop Paths}, and then combine them as the final path-based commonsense knowledge embedding for pair $(a, b)$.

Specifically, we design the message passing mechanism as follows:
\begin{equation}
\text{MSG}^t_v(\mathcal{V}_a) = \text{MSG}^{t-1}_v(\mathcal{V}_a) + \sum_{v \in \mathcal{N}_a} {\text{MSG}^t_e(v)},
\label{eq:msg}
\end{equation}
\begin{equation}
\text{MSG}^t_e(v)= \text{MLP}(\text{MSG}^{t-1}_v(v)),
\end{equation}
where MSG$^0_v(v)$ is initialized as $\bm{F}_v$, \ie, the original node features from \cite{ConceptNet5}. 

In order to update the embedding of each node with all hops on the corresponding paths, given $G_l^{(a,b)}$, we iterate the above process $t=l$ times to obtain the final node embeddings $\bm{T}_l^{(a,b)}$ on each set of \textit{l-Hop Paths} in $G_l^{(a,b)}$. 

To join the information of multiple paths on each graph, we select and aggregate the most important paths. As a simplification, we adopt the  GlobalSortPool operator \cite{zhang2018end} on the embeddings of all nodes in $G_l^{(a,b)}$ as follows:
\begin{equation}
\bm{e}_g^{(a,b),l} = \operatorname{GlobalSortPool}(\bm{T}_l^{(a,b)}),
\end{equation}
which learns to sort all nodes by a specific embedding channel and pool the top K (\eg, K=5) nodes on all embedding channels.

Finally, we aggregate the path embeddings $\{\bm{e}_g^{(a,b),l}\}_{l=1}^{L}$ of different length $l$ through vector concatenation.

\header{Fused integrator.} 
To fuse neighbor- and path-based commonsense knowledge, we inject the neighbor-based knowledge into the path-based knowledge by initializing MSG$^0_v(v)$ in Eq.~\ref{eq:msg} as the element-wise mean between $\bm{e}_{nb}$ from Eq.~\ref{eq:neiE} and the original node features $\bm{F}$ as follows:
\begin{equation}
\text{MSG}^0_v(v) = \operatorname{MEAN}(\bm{F}_v, \bm{e}_{nb}^v), 
\end{equation}
where $v$ can be replaced as $a$ and $b$ for the entity class pair $(a, b)$.

\section{Experiments}
\label{sec:exp}
\subsection{Experimental Settings}
\header{Original whole dataset.} For scene graph generation, we use the Visual Genome dataset \cite{VisualGenome}, a commonly used benchmark for SGG, to train and test our framework. This dataset contains 108,077 images, where the number of classes and relations are 75,729 and 40,480, respectively. However, 92\% of relationships have no more than 10 instances. Therefore, we follow the widely used split strategy on Visual Genome \cite{xu2017scene, gu2019scene, UnbiasedTang_2020} that selects the most frequent 150 object classes and 50 relations as a representative. Besides, we use 70\% images as well as their corresponding entities and relations as the training set, and the other 30\% of images are left out for testing. A 5k validation set is split from the training set for parameter tuning. 

\header{Zero-shot amplified dataset.} 
To further investigate the model's generalization ability in more severe zero-shot settings, we reduce the information that the model can leverage during training by constructing another zero-shot amplified dataset. This is achieved by simply removing images containing less common relations from the training data. As a result, the triplet numbers of the last thirty common relations are halved, while the triplet numbers of the first twenty common relations mostly remain the same. In this way, we exacerbate the difficulty for the model, especially on predicting relations for zero-shot triplets. 
\header{Compared algorithms.} 
We compare \themodel with four baseline methods.
\begin{itemize}
	\item [$\bullet$] \textbf{NeuralMotifs (NM)} \cite{MotifsZellers_2018} is a strong baseline which is widely-compared on Visual Genome for SGG.
	\item [$\bullet$] \textbf{NeuralMotifs with Knowledge Refinement Network (NM+)}   \cite{gu2019scene} is the only existing method that leverages external knowledge for SGG, which is the closest to ours. This method mainly contains two new parts, knowledge refinement and image reconstruction. We add its knowledge refinement part on top of Neural Motifs, which we call NM+. 
	\item [$\bullet$] \textbf{TDE} \cite{UnbiasedTang_2020} is the current state-of-the-art method for scene graph generation. This work is also the first one reporting zero-shot performance on Visual Genome but it does not take effort to improve it. 
	\item [$\bullet$] \textbf{CSK-N} is a baseline based on our framework which makes predictions without visual information. Given a pair of entities, we predict their relation using only the neighbor-based commonsense knowledge embedding.
\end{itemize}

\header{Evaluation metrics.} 
Since the purpose of this work is to improve performance on zero-shot triplets, we follow the relation classification setting for evaluation \cite{UnbiasedTang_2020}. 
Specifically, we use the following two metrics and focus on their evaluations on the zero-shot subset of the whole testing data:
\begin{itemize}
    \item [$\bullet$] \textbf{zR@K}(zero-shot Recall@K): Recall@K is the earliest and the most widely accepted metric in SGG, first used by Lu et al. \cite{lu2016visual}. Since the relation between a pair of entities is not complete, treating it as a retrieval problem is more proper than a classification problem. Here we take the Recall@K on zero-shot subset and shorten it as zR@K.
    \item [$\bullet$] \textbf{ng-zR@K}(zero-shot no-graph-constraint Recall@K):  No-graph-constraint Recall@K is first used by Newell et al. for SGG \cite{newell2018pixels}. It allows a pair of entities to have multiple relations, which significantly improves the recall value. Here we take the ng-R@K on zero-shot subset and shorten it as ng-zR@K. 
\end{itemize}

\subsection{Hyper-parameter setting}
A pre-trained and frozen Faster-RCNN\cite{ren2016faster} equipped with the ResNeXt-101-FPN\cite{he2015deep} backbone is used as the object detector for all models. Batch size and the max iteration number are set as 12 and 50,000 respectively. The learning rate begins with $1.2 \times 10^{-1}$ with a decay rate of $10$ and a stepped scheduler based on the performance stability on the validation set. A Quadro RTX 8000 GPU with 48GB of memory is used for our model training. For path length, here we only use 1 and 2 hop paths ($L=2$) due to the GPU memory limitation. Based on observation from humans' perspective, these short paths are also the more informative ones compared with longer paths. 

\subsection{Performance evaluations}

\begin{table}
\centering
\caption{Zero-shot performance (\%) on the original whole dataset.}
\label{tab:origianl}
\resizebox{1.0\textwidth}{!}{
\begin{tabular}[]{@{}ccccccc@{}}
\toprule
Method & zR@20 & zR@50 & zR@100 & ng-zR@20 & ng-zR@50 & ng-zR@100 \\
\midrule
\textbf{NM} & $13.05\pm 0.06$ & $19.03\pm 0.22$ & $21.98\pm 0.22$ & $15.16\pm 0.49$ & $28.78\pm 0.57$ & $41.52 \pm 0.79$ \\
\textbf{NM+} & $12.35\pm 0.28$ & $18.10\pm 0.13$ & $21.13\pm 0.24$ &
$14.47\pm 0.11$ & $27.93\pm 0.15$ & $40.84\pm 0.28$ \\
\textbf{TDE} & $8.36\pm 0.25$ & $14.35\pm 0.27$ & $18.04\pm 0.46$ & $9.84\pm 0.33$ & $19.28\pm 0.56$ & $28.99\pm 0.44$ \\
\textbf{CSK-N} & $5.95 \pm 0.62$ &$10.12 \pm 0.79$ &$13.05 \pm 0.64$ &$8.15 \pm 0.57$ &$16.79 \pm 0.62$ &$26.19 \pm 1.19$ \\
\textbf{\themodel-N} & $12.73 \pm 0.22$ & $18.88\pm 0.12$ & $21.88\pm 0.11$ & $15.10\pm 0.47$ & $28.73\pm 0.21$ & $41.06\pm 0.26$ \\
\textbf{\themodel-P} & $12.24\pm 0.17$ & $18.12\pm 0.16$ & $21.55\pm 0.39$ & $14.39\pm 0.46$ & $28.90\pm 0.43$ & $40.98\pm 0.45$  \\
\rowcolor{gray!20} \textbf{\themodel-N+P} & $13.42\pm 0.28$ & $19.31\pm 0.27$ & $22.22\pm 0.29$ & $15.54\pm 0.27$ & $29.31\pm 0.27$ & $41.39\pm 0.22$ \\
\bottomrule
\end{tabular}}
\end{table}

\begin{table}
\centering
\caption{Zero-shot performance (\%) on the zero-shot amplified dataset.}
\label{tab:zero-shot_training}
\resizebox{1.0\textwidth}{!}{
\begin{tabular}[]{@{}ccccccc@{}}
\toprule
Method & zR@20 & zR@50 & zR@100 & ng-zR@20 & ng-zR@50 & ng-zR@100  \\
\midrule
\textbf{NM} & $11.98\pm 0.09$ & $17.86\pm 0.13$ & $20.48\pm 0.02$
& $13.98\pm 0.33$ & $27.43\pm 0.72$ & $39.33\pm 0.77$  \\
\textbf{NM+} & $11.82\pm 0.06$ & $17.27\pm 0.34$ & $20.10\pm 0.31$ &
$13.83\pm 0.18$ & $26.92\pm 0.25$ & $38.71\pm 0.32$  \\
\textbf{TDE} & $5.67\pm 0.03$ & $11.08\pm 0.40$ & $14.20\pm 0.22$ & $6.51\pm 0.05$ & $18.61\pm 0.05$ & $32.20\pm 0.47$   \\
\textbf{CSK-N} & $5.65 \pm 0.34$ & $9.55 \pm 0.40$ & $11.74 \pm 0.98$ & $7.35 \pm 0.45$ &$15.04 \pm 0.85$ &$23.91 \pm 1.05$ \\
\textbf{\themodel-N} & $11.79\pm 0.70$ & $17.42\pm 0.88$ & $20.08\pm 1.11$ & $14.14\pm 0.78$ & $26.74\pm 1.38$ & $38.44\pm 1.52$ \\
\textbf{\themodel-P} & $12.17\pm0.84$ & $18.02\pm 1.23$ & $20.58\pm 1.52$ & $14.53\pm 0.92$ & $27.66\pm 1.21$ & $38.86\pm 1.48$  \\
\rowcolor{gray!20} \textbf{\themodel-N+P} & $12.29\pm 0.17$ & $17.85\pm 0.33$ & $20.26\pm 0.46$ & $14.71\pm 0.58$ & $27.67\pm 0.82$ & $39.57\pm 0.54$  \\
\bottomrule
\end{tabular}}
\end{table}

Table \ref{tab:origianl} and Table \ref{tab:zero-shot_training} show performance of three variants of \themodel with different knowledge integrators (\themodel-N, \themodel-P, \themodel-N+P) as well as four baselines on the original and amplified datasets, respectively. The results from baseline methods are reported under their best settings. Using a Bayesian correlated t-Test \cite{test_article}, there is 98\% probability that \themodel-N+P is better than baseline methods. It is shown that on both datasets, our methods \themodel-N+P and \themodel-P surpass all baselines and achieve by far the highest results on both zR@K and ng-zR@K. The performance gains in Table \ref{tab:zero-shot_training} are more significant than Table \ref{tab:origianl}. Such observations directly support the effectiveness of \themodel in zero-shot relation prediction, indicating its superior generalizability as we advocate in this work. 
Note that, consistent with their own report and our analysis in Table \ref{tab:zero_no_zeroperformance}, TDE reaches state-of-the-art on non-zero-shot triplets, but performs rather poor on zero-shot ones.

\begin{figure}[t]
    \centering
    \includegraphics[width=0.5\linewidth]{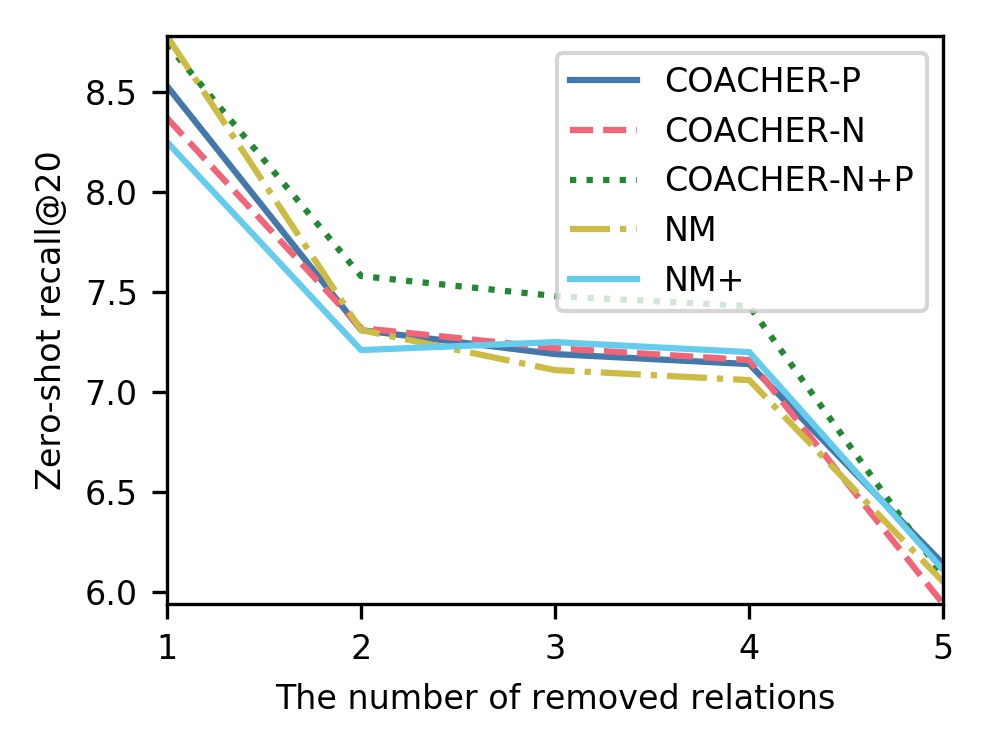}
    \caption{Performance on manipulated testing data (we removed TDE and CSK-N here due to their rather poor performance on zR@20).}
    \label{fig:manipulate}
\end{figure}

\header{Further amplifications on testing data.} 
In order to observe how powerful our model is on harder zero-shot triplets, we further manipulate the testing data by removing zero-shot triplets whose relations are more commonly observed in other triplets in the training data. As is shown in Figure \ref{fig:manipulate}, our proposed method \themodel-N+P shows the least drop compared to other methods as triplets of the top 5 common relations are removed one-by-one from the testing data, which again indicates the advantageous generalization ability of our model. 

\begin{figure}[t]
	\centering
	\subfloat[neighbor-based commonsense helps]{
	    \centering
		\includegraphics[width=0.48\linewidth]{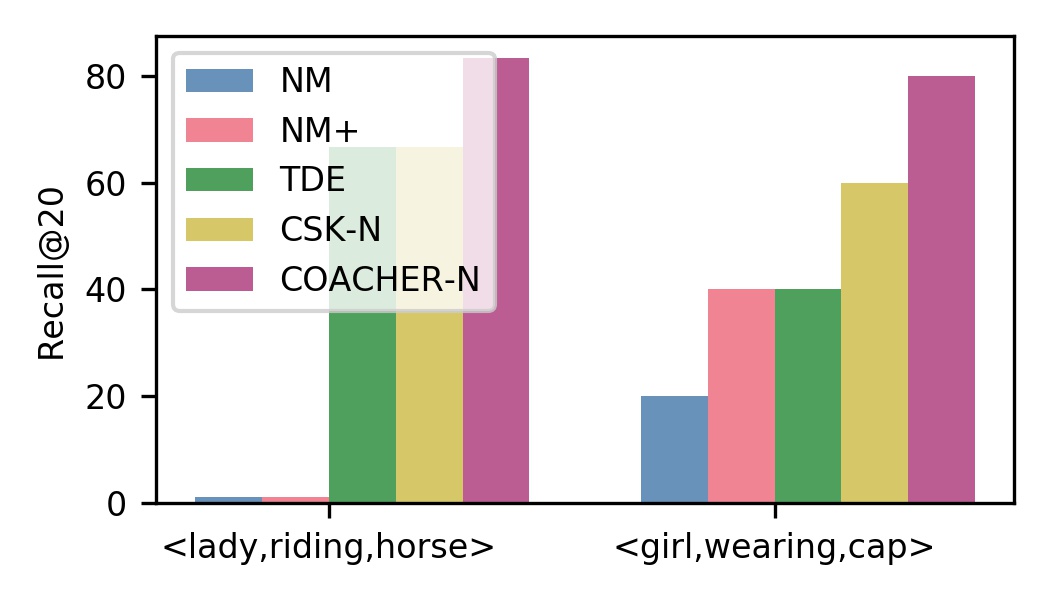}
		\label{fig:case-neighbor}
	}
	\subfloat[path-based commonsense helps]{
	    \centering
		\includegraphics[width=0.48\linewidth]{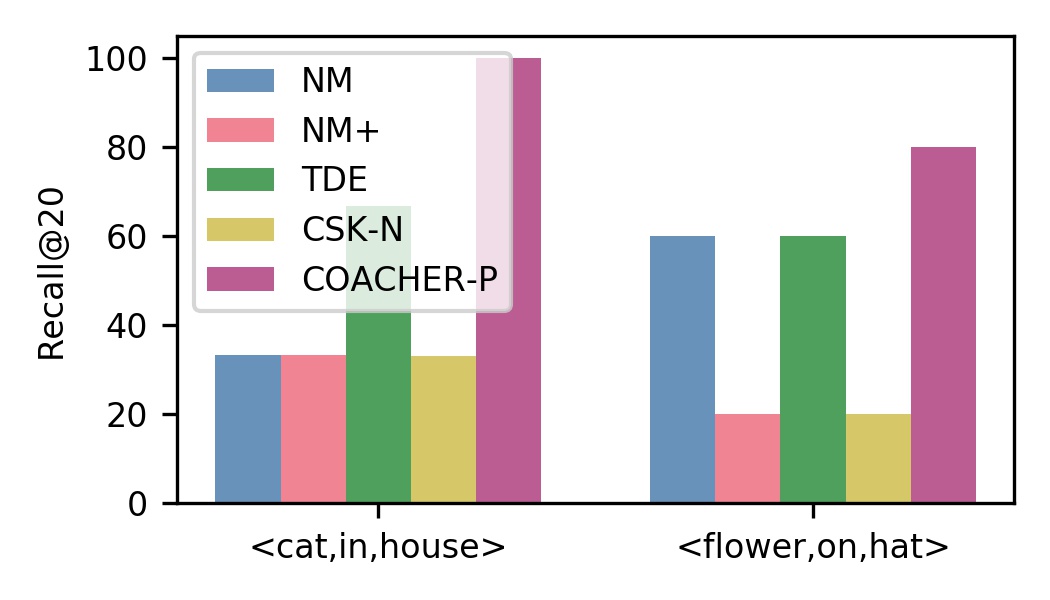}
		\label{fig:case-path}
	}
	\caption{Case studies on integration of neighbor- and path-based commonsense.}
	\label{fig:case-study}
\end{figure}

\subsection{Case studies}
In this subsection, we illustrate the contributions of both neighbor- and path-based commonsense knowledge integrators for SGG with real zero-shot relation prediction examples, shown in Figure \ref{fig:case-study}.
\noindent \header{Neighbor.} 
The contribution of neighbor-based commonsense mainly comes from the semantic similarity implied by the neighborhood overlap between entities. As is shown in Figure \ref{fig:case-neighbor}, although triplet $\langle$\textit{lady, riding, horse}$\rangle$ has never appeared in the training data, several similar triplets such as $\langle$\textit{man, riding, horse}$\rangle$ and $\langle$\textit{person, riding, horse}$\rangle$ are observed in the training data, where their occurrences reach 194 times and 83 times respectively. Estimated from the nodes neighborhood similarity in ConceptNet, the semantic meaning between \textit{lady}, \textit{man} and \textit{person} are similar to each other. \themodel can leverage this external knowledge to improve prediction performance on zero-shot triplets. Similarly, in another case of $\langle$\textit{girl, wearing, cap}$\rangle$, where \textit{cap} is semantically similar with \textit{hat}, since $\langle$\textit{girl, wearing, hat}$\rangle$ is commonly observed in the training data, \themodel can use the neighbor-based commonsense knowledge to make better relation predictions on zero-shot triplets. In contrast, other baseline methods fail in both cases because the generalizable semantic knowledge is hard to directly learn from limited visual information. 
\noindent \header{Path.} 
The contribution of path-based commonsense mainly comes from the relations implied by \textit{MidPaths} between entities. As shown in Figure \ref{fig:case-path}, for the triplet $\langle$\textit{flower, on, hat}$\rangle$, we can find an inductive path $\langle$\textit{flower, RelatedTo, decoration, UsedFor, hat}$\rangle$, which promotes the prediction of relation $on$. Similarly, given the pair $(cat, house)$, multiple paths like $\langle$\textit{cat, AtLocation, home, RelatedTo, house}$\rangle$, $\langle$\textit{cat, RelatedTo, house}$\rangle$ and $\langle$\textit{cat, AtLocation, apartment, Antonym, house}$\rangle$ can be found. All of them are inductive towards the correct relation of $in$. 

\section{Related Work}
\label{sec:related}

\subsection{Scene Graph Generation}
Scene graph generation (SGG) has been widely investigated over the last decade due to its potential benefit for various visual reasoning tasks. Lu et al. \cite{lu2016visual} train visual models for entities (e.g. ``man" and ``bicycle") and predicates (``riding" and ``pushing") individually, and then combine them for the prediction of multiple relations. Xu et al. \cite{xu2017scene} propose a joint inference model to iteratively improve predictions on entities and relations via message passing. Liao \cite{YANG201715} further integrate physical constraints between entities to extract support relations. Importantly, Zellers et al. \cite{MotifsZellers_2018} raise the bias problem into attention that entity labels are highly predictive of relation labels and give a strong baseline. Recently, Tang et al. \cite{UnbiasedTang_2020} present a general framework for unbiased SGG based on causal inference, which performs as the current state-of-the-art SGG method.

All methods above do not leverage external knowledge. The only exception is Gu et al. \cite{gu2019scene}, which adds a knowledge refinement module into the SGG pipeline to leverage external knowledge. It is indeed the closest work to ours and compared as a major baseline. However, they do not consider the zero-shot relation prediction problem and their integration of external knowledge is rather coarse and limited compared with ours.

\subsection{External Knowledge Enhanced Deep Neural Networks}
Knowledge Bases (KBs) can be used as an external knowledge to improve various down-stream tasks. The natural language processing and computer vision communities have proposed several ways to benefit deep neural networks from KBs. 
The most straightforward and efficient way is to represent external knowledge as an embedding, and then combine it with other features to improve model's performance. For example, in order to incorporate external knowledge to answer open-domain visual questions with dynamic memory networks, Li et al. \cite{li2017incorporating, Wu_2018} extract the most informative knowledge and feed them into the neural network after embedding the candidate knowledge into a continuous feature space to enhance QA task. The graph community also utilizes external knowledge to enhance graph learning \cite{yang2020thesis, yang2020heterogeneous, yang2020multisage, yang2020taxogan, yang2019condgen}. Besides, external knowledge can be added to loss function or perform as a regularization in the training process. For example, Yu et al. \cite{Yu_2017_ICCV} obtain linguistic knowledge by mining from internal training annotations as well as external knowledge from publicly available text. 

However, external knowledge has hardly been utilized in SGG, mainly because the direct entity-level embeddings can hardly help in object detection, whereas what kind of embeddings are helpful in what kind of relation prediction has remained unknown before our exploration. 

\section{Conclusion}
\label{sec:conclu}
Scene graph generation has been intensively studied recently due to its potential benefit in various downstream visual tasks. In this work, we focus on the key challenge of SGG, \ie, relation prediction on zero-shot triplets. Inspired by the natural ability of human beings to predict zero-shot relations from learned commonsense knowledge, we design integrators to leverage neighbor- and path-based commonsense from ConceptNet. We demonstrate the effectiveness of our proposed \themodel through extensive quantitative and qualitative experiments on the most widely used benchmark dataset of Visual Genome. 

For future works, more in-depth experiments can be done to study the external knowledge graphs for SGG. Although the current ConceptNet is comprehensive enough to cover all entities detected from Visual Genome images, the relations it models are more from the factual perspectives, such as \textit{has property}, \textit{synonym}, \textit{part of}, whereas the relations in scene graphs are more from the actional perspectives, such as the spatial or dynamical interactions among entities. One promising direction based on this study is to construct a scene-oriented commonsense knowledge graph specifically for visual tasks, while the downstream training process can further refine the graph construction. In this way, the gap between these two isolated communities, \ie, visual reasoning and knowledge extraction, can be bridged and potentially enhance each other. 

\bibliographystyle{splncs04}


%

\end{document}